# Conducting Feasibility Studies for Knowledge Based Systems

John Kingston
Joseph Bell Centre for Forensic Statistics and Legal Reasoning
University of Edinburgh
J.Kingston@ed.ac.uk
www.josephbell.org

Abstract

This paper describes how to carry out a feasibility study for a potential knowledge based system application. It discusses factors to be considered under three headings: the business case, the technical feasibility, and stakeholder issues. It concludes with a case study of a feasibility study for a KBS to guide surgeons in diagnosis and treatment of thyroid conditions.

## 1. Introduction

Many years of experience have demonstrated that knowledge based systems (KBS) are one of the most effective methods of managing knowledge in organisations – if they are applied in appropriate areas and to appropriate tasks. The identification of appropriate tasks and areas is therefore critical – and yet little has been published on this subject, despite renewed interest in the area from knowledge management practitioners. The purpose of this paper is to outline an approach to conducting feasibility studies for knowledge based systems.[1]

There are three major aspects to consider when carrying out such a feasibility study: the business case; technical feasibility; and project feasibility (i.e. involvement and commitment of the various stakeholders). These will be considered in turn. The paper will then conclude with a case study, illustrating these principles being put into practice.

## 2. Feasibility Studies: The Business Case

If a KBS is not expected to bring business benefits, then there is no point in an organisation investing in its development, so the business case must be part of any feasibility study. That much is obvious; what is less obvious is the types of business benefit that a knowledge based system can bring to an organisation.

[1] Acknowledgements are due to the following: Ian Filby (for general knowledge engineering contributions), Knox Haggie (for the case study), Robert Inder (who coined the "telephone test"), Ann Macintosh (for managing and marketing the training course that drove the development of these ideas) and Neil Molony (domain expert for the case study). The case study was supported by EPSRC grant number GR/R60348/01, "Master's Training Package in Knowledge Management and Knowledge Engineering".

The most obvious business benefit is increased productivity, which KBS systems may deliver by reducing the time taken to perform a problem solving task. However, this is rarely the initial motivation for building a knowledge based system; the reasons are normally to do with the need for a knowledge management solution – that is, some operation within the organisation requires expertise, and the expertise is either not available often enough, or not exercised fully. The most common problem with expertise is that it is not available widely enough. The experts may be simply too busy to answer all the queries which require their expertise; alternatively, the experts may be frequently employed on routine cases that do not optimise the use of their scarce expertise. A good example of the latter arose within Ferranti several years ago, when Alan Pridder, one of their staff was tasked with analysing core dumps from military software that had crashed. He became so fed up with poring over mountains of printout, only to find out that someone had kicked the plug out again, that he threatened to resign unless Ferranti did something about it. Their solution was to build a knowledge-based system based on his knowledge that could identify the most common causes of core dumps, thus leaving him with the more interesting cases; the process of having his knowledge elicited was also considered to be an interesting diversion. The resulting system, known as APRES (the Alan Pridder Replacement Expert System) was sufficiently successful that he was still working for Ferranti several years later.

There are also situations where the best expertise is not applied to the problem, usually because of time restrictions; a KBS can provide support in making the best decision. A good example of this is American Express' Authorizer's Assistant system [1], which helps to decide whether transactions can be charged to an American Express card. This is necessary because American Express is a charge card, not a credit card, so the effective credit limit varies according to an individual's credit history, use of the card, and several other factors, rather than being a fixed limit. Transactions on the borderline of acceptability are processed by "authorizers" whose task is to discuss the transaction with a retailer by telephone, look up the customer's credit records, and then make a decision, all in about 90 seconds. The transaction may be granted, rejected, or the card may even be destroyed by the retailer. American Express noticed that some of their authorizers performed much better than others, and so decided to implement a KBS to make the knowledge of the best authorizer available to all others. The project was a major undertaking, but when successfully completed, it saved far more money from the improvements in decision making than from the small reduction in the time required to process each authorisation.

Other cases where improvements in decision making are required include cases where 'experts' disagree (in which case the most senior expert or manager may want to use the KBS to enforce the approach s/he considers best), and cases where there is no real expert. An example of the latter can be found in a system developed by a telecommunications company for diagnosing faults in a new switching system; the system did not exist beyond its paper specifications, so there was no-one with any expertise in diagnosing it! The company's solution to this was to build a model-based reasoning system that could reason from first principles.

There are also further business benefits that may arise from developing a KBS. These may include training of users when they ask for explanations of the system's decisions; it has been shown that providing training to someone when they need to know the answer is a very effective training technique. Management information can also be derived from the workings of the KBS. The organisation may obtain a profile as a user of high technology. One advantage that should definitely not be overlooked is that when an expert is getting close to retirement; the KBS can act as an archive of some or all of the expertise. This approach was used by Campbell's when the expert who diagnosed faults in their giant soup cookers came up to retirement; the expert was described as "slightly bemused that his life's experience had been encapsulated in about a hundred rules".

It is sensible to perform a full cost/benefit analysis, taking into account costs of staff training (to develop the KBS and to use it), hardware and software, and KBS maintenance. As an example, consider ICL's Advanced Coating Plant Advisor, a system for diagnosing faults and advising on recovery of a particular manufacturing plant. In a DTI-sponsored study of this system [2], it was determined that the system had cost £30K to develop (6 man months at a notional rate of £50K per man year, plus £5K for hardware and software) plus an annual charge for knowledge base maintenance of about £5K (1 – 3 man days per month). The benefits were in reduced downtime of the plant (the plant was now online for 95% of the time rather than 92.5%), saving £100K per year; there were also far fewer calls on the expert, cutting his workload by about 80% (equivalent to £50K x 0.8 = £40K per year). So the system paid for itself in 3 months – and rolling out the system to other plants would multiply the benefits.

## 2.1  Organisational feasibility

For a successful feasibility study, it is not sufficient to establish that a KBS could bring benefits to an organisation; it is also important to ensure that the system fits in with the organisation's current or future ways of working. If this criterion is not fulfilled, the system is unlikely to be used after a short period of time.

The most important requirement is to determine how much organisational change will be required in order for the system to be used. It is inevitable that the introduction of a KBS will bring about some organisational changes -- typically some authority will be devolved from the experts to more junior staff. Techniques for handling changes like this are described in the section on Project Feasibility below. However, a KBS that requires major changes in authority or structure is unlikely to be used unless these changes are being carried out independently. A good example of this comes from one of the UK's savings banks, which set up an AI group in the 1980s. The AI group asked for suggested KBS applications from staff, and spent a lot of effort on making a good choice. The final choice was to build a KBS to support the task of mortgage application assessment. Technically and commercially, this was a good decision -- other financial institutions have successfully built KBS to address the same task. However, once the system was built, tested, and demonstrated (successfully), it became clear that the system would be most useful if used by staff in the bank's branches to make good mortgage lending decisions. This would require the bank to devolve its mortgage processing from 6 regional centres to 16,000 branches - a major organisational change which would require considerable redeployment of staff within the regional centres. As a result, the system was quietly shelved.

Another organisational issue is whether the task will continue to be performed. It's reasonably obvious that, if a piece of machinery is obsolete and will soon be phased out, it's pointless to build a KBS to diagnose faults in that machine. However, the longevity of some organisational roles and functions is sometimes less obvious. In practice, it's often easy to spot tasks that *will* continue to be performed, because they deal with the core of the business, or because there will always be a need for these tasks; in other cases, it's worth making enquiries among management if this task is expected to continue for 3-5 years, which is the typical lifetime of an (unmaintained) KBS.

One enquiry that is often useful is "Have you tried any other solutions to this problem? If so, why did they fail?" This is a good way of finding out about any organisational resistance to restructuring or to new technology, which may have a significant effect on the feasibility of the system. It's also possible that other automated solutions have been tried; enquiring about the reasons for the failure of these can provide illuminating technical information. For example, Barclaycard's Fraudwatch system, which monitors Barclaycard transactions to detect spending patterns indicative of possible fraudulent use, was originally implemented using a non-AI computing approach. The reason for failure of this system was that the pattern matching algorithm was not specific enough; with about 100,000 cards being used every day, this system would identify up to 1,000 cards with possible fraudulent transactions, which was far too many for Barclaycard to follow up. The current Fraudwatch system identifies far fewer cards, allowing Barclaycard to telephone many of the card holders and ask if the card is indeed being used fraudulently; if so, the card is cancelled immediately. This system identified 11 frauds in its first 7 days of use, saving Barclaycard an average of £400 per card.

Once a KBS is installed, the effects of *knowledge transfer* and *knowledge seepage* may occur. Knowledge transfer occurs when the KBS has an explanation facility which has a training effect upon the users; the users eventually learn all the knowledge embodied within the system. This effect was observed in the American Express Authorizer's Assistant system, where the use of the explanation facility by new users was monitored. At first, the users accepted the system's recommendations with little interest in the explanations. After a while, they began to look at the explanations frequently; after some more time, they ceased to look at the explanations, having presumably learned everything that the explanations could tell them. It's possible that users may cease to use the system at this stage. A solution to this problem (if it is a problem) is to build a KBS that supplies other benefits of automation; the Authorizer's Assistant, for example, performs fast pattern matching on a database of credit records.

Knowledge seepage occurs when all human expertise in the area is gradually lost as the experts and users become dependent upon the system. This is most frequently encountered in AI systems with adaptive capabilities that update their own knowledge (e.g. neural networks), but may also occur with highly complex KBS. This may be a significant risk to the organisation, particularly in a commercial climate where reorganisations are frequent and far-reaching. Feasibility studies should therefore use the technique of identifying "risk factors" and assessing the impact on the project if these factors should change. For example, the departure of a particular expert who has supplied knowledge for a KBS might be of medium likelihood, but have only a low impact on the project, because knowledge in this domain is very stable. It is wise to build in contingencies to the project plan if there are several risks with both medium/high likelihood and medium/high impact.

## 3. Technical feasibility

### 3.1 Task & knowledge

When assessing the technical feasibility of a proposed system there are various issues to consider. The key one – indeed, this is often the first question asked in a good feasibility study – is the type of task being tackled. KBS have been used to perform a variety of knowledge-based tasks, such as classification, monitoring, diagnosis, assessment, prediction, planning, design, configuration and control tasks. If the task type is not in this list, it is worth asking if it is not more suited to being implemented using non-KBS techniques; for example, a task that primarily involves correlation is better suited to a statistical package than to a KBS.

The form of knowledge is also important in technical feasibility. If the reasoning involved is primarily symbolic reasoning based on concepts, objects or states, then a KBS should be suitable. If there is a significant requirement for calculation, based on numerical data; or a requirement for geometric reasoning, based on graphical data; or (worst of all) a requirement for perceptual input, based on textures, shapes, photographs or facial expressions, then it will be difficult to program a KBS to perform all the necessary operations. Alternative approaches to consider might include CAD packages or computer-based training for humans.

It is often obvious to a knowledge engineer when perceptual input is required, because textbooks or training materials will contain many photographs. However, a good heuristic to determine if any non-symbolic knowledge is required is the "telephone test". It requires the knowledge engineer to ask the expert if, in an emergency, the solution to the problem could be described over the telephone. If the answer is "No", or "It would take a very long time", it's likely that non-symbolic knowledge is involved.

There are some types of knowledge that are definitely suitable for a KBS, and less suitable for other approaches. If the knowledge contains procedures, regulations or heuristics in the form of condition-action statements (If A is true then do B), a taxonomic hierarchy, or a set of alternatives which need to be searched through, then knowledge-based systems are probably the most suitable technology for automating this task. Also, if there is any uncertainty about the knowledge (either knowledge which has confidence factors attached to it, or knowledge which is assumed to be true based on continued belief in other knowledge) then KBS technology has techniques for representing this uncertainty that other technologies do not explicitly support.

A feature that KBS are known to provide well is providing explanations. The explicit representation of knowledge in modular units (i.e. rules or objects) allows the knowledge engineer to attach explanations to individual rules or objects. Explanations are useful both for checking the accuracy of the system's decisions and, as described above, for providing on-the-job training. From a commercial viewpoint, the ability to provide good explanations is one of the most useful features of KBS technology.

Another point which should be included in the feasibility study is to make sure that the knowledge in the KBS is verifiable. In other words, there needs to be an agreed way of checking that the knowledge is correct. This can present quite a problem, for if there is only one expert in a task, who is to say whether the knowledge provided is correct or not? In practice, this is rarely a major problem in the commercial world; perhaps this is because of a greater emphasis on knowledge that achieves the correct result than on knowledge that is provably correct. A knowledge engineer should make sure that the manager funding the project either agrees that the expert's opinions should be considered to be correct, or supplies an alternative "knowledge standard" against which checking can take place.

While discussing verifiability of knowledge with the appropriate manager, it's a good idea to continue to determine what proportion of the task the system should tackle. In other words, when is the project considered to be finished? If this is not specified at the outset, then it's common to find all sorts of extra features or knowledge coverage being requested; if it is specified at the beginning, the knowledge engineer has a clear definition of a successful system. The chief difficulty is that early knowledge acquisition often reveals information about the task and its complexity that affects the definition of success considerably. It's therefore wise to do one or two knowledge acquisition sessions before settling down the definition of a successful system.

## 3.2 Application complexity

Having looked at the task and knowledge to decide whether this problem is feasible for a KBS solution, it's also important to look at how complex the proposed KBS solution would be, for these systems vary widely in levels of complexity; the amount of effort required to implement a commercial KBS can vary from a few weeks to many man years.

A good heuristic for initial estimation of the task complexity is based on the task type. Some task types (principally diagnosis and assessment) are well understood and underlie many KBS applications; others (e.g. planning, design and control) only support a few applications, which are typically complex systems. Task types can be divided into *analytic* tasks such as diagnosis, classification, monitoring and assessment (where there is a finite number of solutions) and *synthetic* tasks such as planning, scheduling, configuration and design (where there is a theoretically infinite number of solutions); the knowledge engineer's heuristic is that analytic tasks are typically less complex than synthetic tasks. This is only a heuristic, however; compare MYCIN, CASNET and INTERNIST, which all have the same task type (diagnosis) and the same general domain (medicine), but have very different levels of complexity.

Another feature worth checking to determine the complexity of the task is the time required by experts to do it. Opinions vary on this, but ideally the expert should take between 3 minutes and 1 hour to solve the problem. If the expert takes less than 3 minutes, then it will be difficult to build a KBS that accepts a meaningful amount of input and solves the problem as quickly as the expert; American Express managed it, but that was a million-dollar project. If the expert takes more than an hour to solve the problem, then it may be that the problem has many sub-components, and it would be better to begin by implementing a KBS to tackle one of the component tasks.

The biggest potential time sinks in any project are the interfaces. Interfaces may take up to 80% of the code for the whole system, and will take up 10-50% of the project time. If the application requires several interfaces to other systems (e.g. databases), or if an impressive-looking user interface is required, then the knowledge engineer should make allowances in the project budget for 30-50% of the effort to be spent on interfaces. If the client will be content with an embedded system or a *simple* mouse-and-menu interface, then 10-15% is more realistic.

Another factor that greatly affects complexity is criticality. If it is critical that the system's answers are always correct and provide 100% coverage of the domain (for reasons of safety, or because there is the risk of significant loss of money, etc.) then the development process will require much more effort. The "80/20" rule states that building a system with 80% coverage (and 100% accuracy) takes 20% of the time required to build a system with 100% coverage, so it's sensible to aim for 80% coverage if that is acceptable.

Looking at the knowledge again, it isn't sufficient to determine if the knowledge is symbolic or not. Certain types of knowledge may be represented as concepts, objects or states, but may still be very complex for a KBS to reason with. These include temporal knowledge (time-based orderings or time restrictions), spatial knowledge (e.g. the location of a desk relative to a door in an office layout problem), cause-effect reasoning at a 'deep' level (e.g. encoding the laws of physics and using them to make predictions), or a requirement to process real-time data inputs. There are existing KBS systems which work with each of these types of knowledge, so none of them make a KBS infeasible, but they do increase the complexity of the task. However, if common-sense reasoning is required, then a KBS is likely to have severe problems. Intelligence and common sense are not the same thing, as many parents of intelligent children will tell you, and without a huge "life knowledge" database, which is beyond the current scope of KBS technology, KBS cannot perform common sense reasoning.

The final factor to consider is validation; that is, judging if the system gives the correct answers based on the knowledge put into it. This can be difficult to do in a live situation, because the correctness of some systems (e.g. loan advisory systems) cannot be judged by their results until years later. The accepted practice is to devise a test suite based on past cases of problem solving, and to make sure that the KBS produces the correct answer for each of these. It's wise to ask the client to agree to the adequacy of this test suite, since conformance to the test suite will be a significant factor in defining a successful system.

Just as with the business feasibility, there will be risk factors that might affect the technical feasibility of the KBS. These should be identified, with high-likelihood high-impact risks being noted in the feasibility study, and contingency plans made.

## 4. Stakeholder issues

Stakeholder issues – getting involvement and commitment from all parties involved with the system – is often considered the last important area in a feasibility study. In practice, however, more systems fail to be used because of project factors than for any business or technical reasons. The stakeholders involved will be management, users, developers, and the experts whose knowledge is being used in the system; these will be considered in turn.

### 4.1 Management

Management must agree that the feasibility study is adequate, must be willing to fund the system and make key personnel available throughout its development, and should support any organisational changes required to introduce the system. Some organisational change is inevitable, but if the organisational changes are small (e.g. devolving authority for routine problem solving from the expert to junior staff or an autonomous system) and well-justified (explanations of the KBS' reasoning can help here), then the changes can be made easier by allowing the expert a monitoring role. If the system is being introduced as part of a deliberate organisational change, it is up to management to ensure an adequate role (and adequate support) for the KBS in the new structure.

### 4.2 Users

Users must be willing and able to use the system. The ability to use the system can be ensured through training - typically a day's training for one or two people from each user department is sufficient, though training for all users is (of course) ideal. Willingness to use the system is sometimes more difficult to create; giving the users increased authority via the system may be a sufficient incentive, but it's most important that the users understand the justification for the system. An example can be found in a system built for police patrol officers in Ottawa to help with identification of patterns in residential burglaries [3]. The system required patrol officers to spend more time collecting data than they had done previously, so there was a risk that they might not use it. The knowledge engineers handled this by giving the patrol officers slightly more authority (they were permitted to close some cases, rather than referring everything to detectives) and also by giving a 2-hour presentation to all patrol officers in which 90 minutes was spent explaining the justification for the system, and half an hour on how to use the system. The knowledge engineers also demonstrated their own commitment to the project by giving these presentations at 5am, which was the only time when significant numbers of patrol officers could be spared from policing duties!

### 4.3 Developers

The KBS developers need to know how to do knowledge acquisition, how to build KBS in a structured manner, and how to use the chosen programming tool. The best way to deal with this is to choose a tool that the developers already know well. Any deficiencies in developers' abilities can be remedied by sending them on training courses, which should be built into the cost/benefit analysis of the project.

### 4.4 Experts

For the expert, the issues that might arise are as follows:

- The expert may be senior to the knowledge engineer;
- The expert may be uncomfortable with describing his job verbally;
- The expert may be too busy to spend time with the knowledge engineer;
- The expert may perceive the system as a threat to his job security;
- The 'expert' is not really an expert at all.

The first two issues can be handled by starting knowledge acquisition with techniques that the expert is comfortable with (e.g. interviews) rather than techniques that are most beneficial to the knowledge engineer (e.g. card sorting). If the expert is very busy, it is important to ensure that knowledge is available from some alternate source, whether it be lesser experts, manuals, previous cases of problem solving, so that meetings with the expert can be kept to a minimum. The 'threat' issue can be handled by building an "80/20" system, thus retaining an active role for the expert; by giving the expert authority over maintenance or knowledge updates to the system; or by choosing an expert who is about to retire, when this is no longer an issue. The issue of non-expert experts is a difficult one, because there is a significant risk that the knowledge engineer will make himself very unpopular by exposing this; some quiet words with an sympathetic senior figure in the client organisation might result in a change of expert, or a change of focus for the KBS.

### 4.5 Other project issues

An issue that is of great significance for KBS is maintenance. Although the knowledge within KBS is often easier to maintain than the code in many other computing systems, many knowledge-based applications require the knowledge to be updated much more frequently. For this reason, systems that have fast-changing knowledge (such as help desks for rapidly changing products such as computer hardware or software) are often based on case-based reasoning, which combines aspects of knowledge-based and adaptive technologies. For a KBS, the knowledge engineer should ensure that the feasibility study considers knowledge maintenance, and encourages management to select someone capable of knowledge maintenance. A good solution is to give the expert himself enough training that he is able to understand the knowledge base himself; he can then take on responsibility for knowledge maintenance, even if he does not do the actual programming.

## 5. Case Study: An Internet-based Clinical Protocol

The preparation of a feasibility study will be illustrated by referring to a case study of the development of systems to support doctors in using clinical protocols via the Internet. The protocols are for treatment of thyroid problems, and the system for which this study was prepared has recently been developed to prototype level for New Cross Hospital in Wolverhampton. The technical approach used is similar to that described in [4].

Medicine was one of the earliest application areas for knowledge-based systems: the MYCIN system [5], which recommended antibiotics based on clinical data was the first commercially viable KBS to be produced. Since then, KBS have been introduced throughout the medical field; today, systems can be found in routine use in areas such as managing ventilators in ICUs [6], hepatitis serology [7], clinical event monitoring (based on the Arden syntax) [8; 9], diagnosis of dysmorphic syndromes [10], CSF interpretation [11], and other areas [12].

All these KBS examples are from the practice of medicine or anaesthesia; in many surgical specialties, sharing and re-use of knowledge in many medical fields is still limited to the dissemination of experiences and distilled knowledge by the traditional approaches of seminars, journals, and practical training. However, there is a growing trend to promote "best practice" within a specialised area through the use of *clinical protocols*. The idea is to provide guidelines based on strong scientific evidence. Protocols at present exist as intra-department guidelines for the management of clinical situations; where they exist, it is expected that they will normally be adhered to unless there is a good counter-argument. They are principally used to benefit sub-consultant grades. There are as yet only the beginnings of formalised nationally agreed protocols (e.g. those published by the Scottish Intercollegiate Guidelines Network). They are usually printed sheets rather than computer-based, with copious references to the published clinical studies that justify each recommendation.

This feasibility studies considers development of a system that will assist clinicians in following clinical protocols for the diagnosis and treatment of thyroid-related conditions. The systems will reason about the weight of available evidence (in the form of published clinical trials), and will also provide access to relevant publications if requested. The expected users are surgeons who would normally make use of a written protocol, and who will make the final decision on whether to follow the system's recommendations.

## 5.1 Business Case

The heart of the business case lies in improving decision making by automating the protocol, making it easier for surgeons to follow (or seek justification of) its recommendations. This can achieved by encoding details of the published clinical studies and reports that justify each step, together with a measure of the reliability (see [13]) of each study. These measures could then be combined to produce qualified recommendations.

The system would also have associated benefits in providing on-the-job training. If the system is regularly updated (and the beauty of an Internet-based system is that it only needs updating on one computer), then all users of the system will be made aware of new studies in the field which support or supersede old studies – or at least, of the effects of those studies on decision making.

The analysis of costs and benefits is an important issue for any IT system. The financial benefits obtained by obtaining *faster* cures can be estimated in terms of savings in salary, time and associated costs. If a new out-patients appointment costs £70, and a review £50, there are huge savings to be made by minimising reviews and reaching a decision at the first clinic visit. To halve the review appointments in a single department would save £600,000 per year; alternatively the routine new patient waiting time for an appointment could reduce from 8-10 weeks to around 2 weeks if referral rate remained steady. The positive effect of this on patient satisfaction should be significant.

There are also possibilities of *more effective cures* or *longer lasting* cures, by reducing erroneous (or, more likely, sub-optimal) decisions made by junior clinicians. The financial benefits of this are hard to quantify, but should manifest in fewer repeat visits, a reduction in exposure to claims for financial damages, and further increases in patient satisfaction. The potential benefits of a system like this are therefore greater than the "bottom line" figure of £600,000 p.a. would suggest.

Balanced against these expected benefits, we must consider the investment required. Based on the experience of similar KBS projects, it is estimated that a fully functional prototype system would take about six months of effort to complete, with a further three months of effort for testing, revision, installation and training. This translates roughly to £45,000 of development costs (at a notional rate of £60,000 per man year). In addition, there will be hardware and software costs to cover. For software, the ideal software package, chosen after a review of available packages (see [14]) is CORVID from Exsys, whose list price is $10,000 for a development version plus $6,000 for a server-based runtime license. Hardware costs and maintenance could be around £3,000, with replacement every three years. Also, there will be maintenance of the knowledge base to read clinical studies and keep them up to date; allowing two or three days per month, this is estimated at £7,500 per year. Altogether, these figures produce an initial required investment of around £55,000 plus a total annual maintenance cost of approximately £9,000. The system would therefore pay for itself in less than 2 months if a 50% reduction in review appointments could be achieved across a whole department; it is sensible, however, to roll out a prototype first to see how achievable this 50% reduction is.

There are some further organisational considerations that should be considered before the business case for this system is declared to be sound. The system doesn't require any change to organisational responsibilities, unlike some pioneering expert systems that aimed to replace doctors rather than supporting them, although it may result in in junior consultants being able to take more responsibility in the decision process. The task of following surgical protocols is highly unlikely to be phased out in the near future. And for a relatively short IT project such as this one, the risk of funding being cut before the end of the project is comparatively low.

### 5.2 Technical issues

The technical issues affecting these systems are as follows:

- Task type: It is clear that clinical protocols are used to carry out a diagnostic task, so KBS technology looks like a suitable approach.

- Telephone test: Protocols are initially drafted in textual form by a consultant using a specialised vocabulary, and the decision-making process seems to be couched entirely in symbolic terms. This suggests that a KBS approach is appropriate for the task. Some terms do refer to the classification of patterns seen through a microscope, but these descriptions ("sheets", "follicular cells", etc) are manipulated entirely symbolically as far as the protocol is concerned. In other words, the pattern recognition (performed in the hospital's Cytology laboratory) is clearly outside the scope of the system.

- Uncertain knowledge: the knowledge appears to be largely procedural with a need to handle some uncertainty at the decision points.

- Safety-critical: Clearly the task of effective use of protocols is a safety-critical task; however, it is impossible to ensure that an expert system is infallible when agreement amongst surgeons on what is the "best" procedure is still being developed. The aim of this application is to represent the best available knowledge, thus improving on the current situation.

- Verifiability: the knowledge can be verified through clinical trials.

- Complexity: there is little or no requirement for representing temporal or spatial information, cause-effect reasoning or in handling real-time inputs.

- Time required: while the diagnostic process may be spread out over a long period (days or weeks) while test results are awaited, the actual problem-solving time for the vast majority of cases seems to fall into the 3 – 60 minutes range deemed acceptable for a KBS. Very occasionally, consultants will meet together to discuss a more complicated case.

- Interfaces: The aim of this project would be to implement a system that runs within an Internet browser, so that it can be used on an intranet or over the Internet. All interfaces will therefore be written in HTML. Providing links to the studies that provide the evidence for decisions is also required; it is planned that hyperlinks to the Medline online abstracting/publications service will suffice. Links to electronic medical records are not planned since there is no agreed format for these at present.

### 5.3 Stakeholder issues

- Management: The 'management' for this project are hospital consultants who will also function as domain experts. All the consultants involved seem very keen to pursue the project.

- Users: will initially be junior doctors working for these consultants, so should be enthused and encouraged by management. The development of a prototype, fielded to a limited number of health care professionals for evaluation purposes, will give prospective users a chance to comment on all aspects of the system; its usability, its content, and its decision-making. It is also hoped that a medical evaluation will be possible, in which some patients are treated according to advice given by the system (and approved by the health care professionals), and the results are evaluated

- Developers: University of Edinburgh staff with experience in programming KBS are available to develop the system.

- Experts: see above.

## 6. Conclusion

This paper has shown how a feasibility study can be developed for a knowledge-based system, focusing on business, technical, and stakeholder-related issues. In each section, it has highlighted important factors to consider and explained why they are important. A case study was also presented that demonstrated the feasibility – with some caveats regarding interfaces, safety criticality and user acceptance – of a knowledge based system to support the use of clinical protocols in the diagnosis and treatment of thyroid conditions.

Since the task of developing a feasibility study is itself a knowledge-based assessment task, further work in this area might focus on a meta-analysis of these various factors. Issues that might be considered are:

- Priority: which of the factors considered above are showstoppers, and which are merely risk factors that can be managed with contingency plans?

- Tradeoffs: e.g. is it worth making sacrifices in technical feasibility to enhance user acceptance of the system?

- Ideals: what are the features of an ideal KBS application domain?

- Extensibility: how many of these factors apply to approaches similar to KBS: case based reasoning, neural networks, other approaches?